\documentclass[conference]{IEEEtran}
\ifCLASSINFOpdf
\else
\fi
\usepackage[cmex10]{amsmath}
\makeatletter
\let\IEEEcaption\@makecaption
\makeatother

\usepackage[font=footnotesize]{subcaption}

\makeatletter
\let\@makecaption\IEEEcaption
\makeatother

\usepackage{graphicx}

\usepackage[utf8]{inputenc}
\usepackage{textcomp}
\usepackage{booktabs}
\usepackage[dvipsnames]{xcolor}
\usepackage{tikz}
\usetikzlibrary{calc,intersections,decorations}
\usepackage{pgfplots}
\pgfplotsset{compat=1.8}

\pgfplotsset{
	axis line style={black}
}

\tikzset{black_neuron/.style={
		draw=black!50,
		shape=circle,
		minimum size=2.5mm,
		inner sep=0,
		smooth,
		font=\tiny
	}
}
\tikzset{red_neuron/.style={
		draw=black!20!red,
		shape=circle,
		minimum size=2.5mm,
		inner sep=0,
		smooth,
		dashed,
		thick,
		font=\tiny
	}
}

\tikzset{red_link/.style={
		draw=black!20!red,
		smooth,
		thick
	}
}

\newcommand{\chromosome}[3]
{%
	\subcaptionbox{#1}
	{
		\begin{minipage}{#2}
			\tiny
			#3
		\end{minipage}
	}
}

\usepackage[%
style=ieee,
bibstyle=numeric,
useprefix=false,
maxbibnames=99,
uniquename=false,
firstinits=true,
isbn=false,
doi=false,
sorting=ayt,
backend=biber,
sortcites=true,
urldate=comp
]{biblatex}

\DeclareSortingScheme{ayt}
{
	\sort{
		\field{presort}
	}
	\sort[final]{
		\field{sortkey}
	}
	\sort{
		\name{sortname}
		\name{author}
		\name{editor}
		\name{translator}
		\field{sorttitle}
		\field{title}
	}
	\sort{
		\field{sortyear}
		\field{year}
	}
	\sort{
		\field{sorttitle}
		\field{title}
	}
}

\DeclareSortingScheme{yat}
{
	\sort{
		\field{presort}
	}
	\sort[final]{
		\field{sortkey}
	}
	\sort{
		\field{sortyear}
		\field{year}
	}
	\sort{
		\name{sortname}
		\name{author}
		\name{editor}
		\name{translator}
		\field{sorttitle}
		\field{title}
	}
	\sort{
		\field{sorttitle}
		\field{title}
	}
}

\DeclareNameAlias{sortname}{last-first}
\DeclareNameAlias{default}{last-first}

\DefineBibliographyStrings{english}{%
	bibliography = {References},
}

\DefineBibliographyStrings{english}{%
urlseen = {Accessed on}%
}

\usepackage[parfill]{parskip}

\renewbibmacro{in:}{}

\newcommand{\red}[1]{\textcolor{Maroon}{\textbf{#1}}}
\newcommand{\green}[1]{\textcolor{ForestGreen}{\textit{#1}}}

\usepackage{chngcntr}
\counterwithout{footnote}{section}

\usepackage[bottom]{footmisc}

\begin{document}
%
\title{Complexity-based speciation and genotype representation for neuroevolution}

\author{%
	\IEEEauthorblockN{Alexander Hadjiivanov}
	\IEEEauthorblockA{School of Computer Science and Engineering\\
		University of New South Wales\\
		Sydney, NSW 2052, Australia \\
		Email: a.hadjiivanov@student.unsw.edu.au}
	\and
	\IEEEauthorblockN{Alan Blair}
	\IEEEauthorblockA{School of Computer Science and Engineering\\
		University of New South Wales\\
		Sydney, NSW 2052, Australia \\
		Email: blair@cse.unsw.edu.au}
	}


%


\maketitle

\begin{abstract}
This paper introduces a speciation principle for neuroevolution where evolving networks are
grouped into species based on the number of hidden neurons,
which is indicative of the complexity of the search space.
This speciation principle is indivisibly coupled with a novel genotype representation which
is characterised by zero genome redundancy, high resilience to bloat, explicit marking of recurrent connections,
as well as an efficient and reproducible stack-based evaluation procedure for networks with arbitrary topology.
Furthermore, the proposed speciation principle is employed in several techniques designed to promote and preserve diversity
within species and in the ecosystem as a whole.
The competitive performance of the proposed framework, named Cortex, is demonstrated through experiments.
A highly customisable software platform which implements the concepts proposed in this study
is also introduced in the hope that it will serve as a useful and reliable tool for experimentation in the field of neuroevolution.
\end{abstract}


%
\IEEEpeerreviewmaketitle

\section{Introduction}\label{sec:intro}
The process of evolving neural networks (neuroevolution; NE) has been the subject of extensive research for more than two decades.
Although early research on NE focused on fixed-topology networks \parencite{Montana_Davis--1989,Wieland--1991,Richards_Moriarty_Miikkulainen--1998},
where only the synaptic weights were evolved, the benefits of evolving the topology as well as the weights have also been recognised \parencite{Angeline_Saunders_Pollack--1993,Gruau--1993,Stanley--2004,Stanley_Mikkulainen--2004,Gauci_Stanley--2010,Risi_Stanley--2012}.
However, the evolution of arbitrary network topologies introduces additional complexity in terms of genome representation,
which is part of the reason why fixed-topology NE has not fallen completely out of favour \parencite{Gomez_Schmidhuber_Miikkulainen--2008}.
This is related to the problem of \textit{competing conventions} \parencite{Schaffer_Whitley_Eshelman--1992,Angeline_Saunders_Pollack--1993},
whereby the same network can be encoded by two or more different genomes, which reduces the effectiveness of the genome representation.
Furthermore, allowing neural networks to evolve arbitrary topologies may render it difficult to match genomes encoding different topologies
when performing crossover \parencite{Stanley--2004}.

An issue of particular concern is that structural mutations in networks (the addition and deletion of neurons and connections)
can initially degrade the performance of the network, even if the mutation is beneficial in the long run \parencite{Stanley_Mikkulainen--2004}.
This problem can be mitigated by grouping networks into \textit{species}, as a result of which networks only compete directly with other networks within their own species
rather than competing against the entire ecosystem.
While \textit{niching} \parencite{Miller_Shaw--1996} as a form of speciation has been used to maintain diversity and prevent premature convergence to a suboptimal solution,
speciation with such explicitly protective function was introduced more recently in the context of the NEAT framework \parencite{Stanley_Miikkulainen--2002}.
In NEAT, speciation depends on the number of disjoint and excess genes \parencite[pp. 109--110]{Stanley_Miikkulainen--2002} and the difference in mean synaptic weight.
The importance of each of these three parameters can be adjusted by setting the corresponding parameter coefficients in the model.
While the ability to adjust the granularity of speciation provides flexibility, choosing suboptimal values for the coefficients may compromise the performance of the model.
Furthermore, although excess and disjoint genes indicate the addition and/or deletion of connections to the genome
(and respectively to the network), this scheme does not directly take into account
the addition or deletion of neurons, which is arguably the most disruptive form of mutation occurring during NE.

In the present study, the above problems are addressed by adopting a genotype representation and speciation framework which facilitates crossover by providing a simple way to
match different network topologies. The genotype is very similar to the phenotype, and therefore the encoding is direct.
This eliminates the need to decode the genotype into a network in order to evaluate its fitness, and allows the output to be computed in an efficient and reproducible way.
Furthermore, the proposed representation is free of redundancies such as non-expressed genes, and thus becomes highly resilient to bloat.
Finally, well-defined speciation based on the complexity of the search space emerges naturally, without the need for any adjustable parameters.

This paper is organised as follows. Section~\ref{sec:previous_work} provides a brief overview of existing research
and the relevant problems addressed in this study.
Section~\ref{sec:cortex} introduces the proposed NE framework, and Section~\ref{sec:experiments} presents the results of
experiments demonstrating some of its advantages. Finally, Section~\ref{sec:conclusion} concludes the paper with a discussion of possible directions for future research.

\section{Previous Work}\label{sec:previous_work}

As outlined in Section~\ref{sec:intro}, although evolving the topology as well as the weights of networks is sometimes problematic, it can be beneficial if designed and implemented properly.
This section gives a brief overview of existing relevant research, with a focus on evolving network topology.

One of the early methods for evolving networks with arbitrary connectivity was the connectivity matrix scheme \parencite{Kitano--1990}.
This method was used, for example, to evolve application-specific controllers with optimal size and structure \parencite{Dasgupta_Mcgregor--1992}.
Although evolving the connectivity proved to be more effective than fixed-topology methods, the number of neurons was essentially limited.
One of the main advantages of evolving the network topology is that it eliminates the need to decide not only the connectivity, but also the number of hidden neurons.
There is no rigorous method to decide \textit{a priori} how many hidden neurons are necessary in order to obtain optimal performance on a given task,
and the difficulty of finding the optimal number of hidden neurons increases proportionally to the complexity of the task.
Note that this problem is not specific to NE and is unrelated to the method used for training the network.

One framework for evolving network topologies which allows for dynamic addition of new neurons is cellular encoding \parencite{Gruau--1993}.
In this framework, networks are represented as graph grammars evolving through a process resembling cell division.
In a formal comparison on the double pole balancing task \parencite{Wieland--1991}, cellular encoding was able to evolve
networks which generalised considerably better than ones with a fixed topology, while being much more compact ($0$--$2$ hidden neurons
versus $10$ for the fixed-topology network) \parencite{Gruau_Whitley_Pyeatt--1996}.
This result demonstrated that a seemingly very difficult task might be solvable by a neural network controller whose structure is much simpler than anticipated by a human
manually designing a controller for the same task.
In this regard, it has also been demonstrated that neural networks with evolved topology are capable of generalising better than fixed-topology networks trained with backpropagation \parencite{Dodd--1990}.

Evolving network topology by adding and deleting neurons is more involved than using a fixed number of neurons, and requires a suitable genome representation.
The NeuroEvolution of Augmenting Topologies (NEAT) framework introduced a genome representation which proved to be highly successful in this respect \parencite{Stanley_Miikkulainen--2002,Stanley--2004}.
In NEAT, the genome is linear and grows gradually with the addition of new neurons and connections \parencite[Fig. 3]{Stanley_Miikkulainen--2002}.
When a new neuron is added, an existing connection is \textit{disabled}, and two new connections to and from the new neuron are added.
The rationale behind this scheme is that it introduces novel behaviour while minimising the impact on the network's fitness.
As different genomes grow, they diverge and gradually become less and less compatible with each other.
To address this issue, NEAT also introduced the concept of \textit{speciation} with the explicit function of protecting innovation.
Specifically, genomes are compared in terms of the number of excess and disjoint genes and the difference in mean synaptic weight.
If the discrepancies are sufficiently large, the genomes are placed in different species in order to prevent unfair competition between networks of different complexity.
These ideas proved to be very successful and largely drove the surge in popularity of the NEAT framework, which has been extended with additional functionality \parencite{Gauci_Stanley--2010,Risi_Stanley--2012}.

The present study proposes a speciation scheme based on the complexity of the phenotype in terms of number of hidden neurons.
This idea is inspired by the relatively recent discovery that several regions in the brain
(notably, the dentate gyrus in the hippocampus) continue to produce \textit{new} neurons throughout
the life of an animal (at least in birds and mammals, including humans). Although at present the function of this sustained
neurogenesis is not clear, there are indications that it might be associated with learning \parencite{Gould_Beylin_Tanapat_Reeves_Shors--1999},
the formation of temporal memories \parencite{Aimone_Wiles_Gage--2006} and the complexity of tasks, the richness of the environment and/or general experience \parencite{Gage--2002,Kempermann--2002}.
In addition, it seems that the relation between neurogenesis and learning is bidirectional (in other words, learning stimulates neurogenesis and vice versa) \parencite{Kempermann--2002}.
Note that the proposed speciation method is not necessarily biologically accurate in the sense of speciation as seen in living creatures
on Earth, which would fall in the realm of artificial life.
Rather, it draws a parallel with the process of generation and removal of neurons in the brain in response to the complexity of the tasks being performed.
The following section presents a complete NE framework named Cortex, which combines the abovementioned concept of speciation
with a novel genotype representation, relevant genetic operations (crossover and mutation), as well as several techniques for promoting and preserving diversity
which make use of speciation.

\section{Cortex Neuroevolution Framework}\label{sec:cortex}

\subsection{Genotype representation}
First, a note about terms and conventions is in order. The terminology used in this study is adopted in an attempt to
avoid the ongoing controversy regarding the rigorous definition of terms such as genome, genotype and phenotype \parencite{Mahner_Kary--1997}.
Specifically, \textbf{population} refers to a group of \textbf{individuals} of the \textbf{same species},
and the collection of all populations is referred to as \textbf{ecosystem}.
In Cortex, the \textbf{genome} is common to all individuals belonging to a species.
In fact, species are \textit{defined} through their respective genomes, and it is up to the individuals to \textit{express} the genome of their species by forming connections,
thus creating the \textbf{genotype}. The genotype consists of three chromosomes representing outward, inward and recurrent connections.
Based on this, we use the term \textit{genotype} representation rather than \textit{genome} representation to emphasise the fact that the genome is an abstract
notion common to many individuals belonging to the same species, while the genotype encodes information about a particular individual.

The genotype representation in Cortex is inspired by the connectivity matrix representation \parencite{Kitano--1990} and the NEAT representation \parencite{Stanley_Miikkulainen--2002}.
Like NEAT, it adopts a strict order for neurons: bias, input, output, hidden. However, the NEAT genome is linear, whereas the genotype in Cortex is two-dimensional.
It is essentially composed of connectivity tables, which are reminiscent of connectivity matrices, with the exception that
only existing connections are marked explicitly and recurrent connections are stored separately.
An example of a genotype with the corresponding phenotype is presented in Fig.~\ref{fig:cortex_genotype}.
There is no redundancy in the genotype since non-existent connections are not expressed.
As a result, the genotype closely resembles the phenotype, similarly to the PDGP representation scheme \parencite{Pujol_Poli--1998}.
Note that this representation minimises the competing conventions problem because as long as neurons have identical activation functions,
they are indistinguishable from each other and cannot themselves participate in distinguishing network topologies.

Although the table of inward connections can be generated from the table of outward connections at runtime just before the network is evaluated,
this will incur a time overhead proportional to the size of the outward connection table.
Therefore, the inward connection table is presented as part of the genotype in order to illustrate the stack-based evaluation procedure in Fig.~\ref{fig:cortex_evaluation}.

\subsection{Speciation}
The number of hidden neurons in a neural network determines the dimensionality of the search space that it can explore.
Therefore, it is a direct indicator of the upper limit of the level of complexity of the task that the network is capable of learning or solving.
Thus, the specific connectivity pattern of the phenotype determines what portion of this space is being explored.
A fully connected network can potentially explore the entire available space, whereas a partially connected one explores only part of it.
However, if a particular problem requires a minimal number of hidden neurons, a network exploring a smaller search space will never be able to find a solution.
The immediate advantage of Cortex in this respect is that speciation is automatic, being based solely on the number of
hidden neurons\footnote{Technically, the \textit{total} number of neurons, but in the current version of Cortex, only hidden neurons can be added or deleted.
Therefore, the number of bias, input and output neurons remains constant for all individuals throughout their life.}.
The granularity of speciation does not depend on any adjustable parameters.

It is noteworthy that the generic version of NEAT supports only the addition of neurons. This leads to the problem that the NEAT genome can
evolve only networks of similar or higher complexity, without any means of \textit{reducing} the complexity of the evolved topologies.
Although an extension referred to as a \textit{pruning phase} was introduced in order to allow NEAT to delete neurons and simplify topologies,
the solution seems \textit{ad hoc} because it is activated when the average complexity reaches a certain pre-set threshold.
In contrast, the ability to delete neurons is built into Cortex, and thus by setting the respective rates of neuron addition and deletion,
it is possible to evolve an ecosystem of networks whose complexity increases, decreases or `spreads out' evenly in both directions.

\subsection{Genetic operations}
Cortex supports the usual genetic operations, such as crossover and different mutation types. Each type of operation is briefly outlined below
together with some principles for promoting diversity within species and in the ecosystem as a whole.

\subsubsection{Crossover}
The genotype representation offers a very direct way to match the topologies of two networks because we know that the neurons are ordered in a particular way.
An example of a crossover procedure is presented in Fig.~\ref{fig:cortex_crossover}.
Mating between individuals from different species is also possible, with excess neurons and their corresponding connections inherited directly from the longer genotype where appropriate.

\begin{figure}
	\centering
	\subcaptionbox{}[0.25\columnwidth]
	{
		\centering
		\begin{tikzpicture}[xscale=0.5,yscale=0.75]

		\foreach \l [count=\i] in {1,2,3,4}
			\node [black_neuron] (i-\i) at (\i - 1,0) {\l};

		\foreach \l [count=\i] in {7,8,9}
			\node [black_neuron] (h-\i) at (\i - 0.5,1) {\l};

		\foreach \l [count=\i] in {5,6}
			\node [black_neuron] (o-\i) at (\i,2) {\l};

		\foreach \l [count=\i] in {1,...,4}
			\draw [<-] (i-\i.south) -- (\i - 1,-0.5) {};

		\foreach \l [count=\i] in {1,2}
			\draw [->] (o-\i.north) -- (\i,2.5) {};

		\draw[->] (i-1) to (h-1);
		\draw[->] (i-1) to (h-2);
		\draw[->] (i-1) to (o-2);
		\draw[->] (i-2) to (h-1);
		\draw[->] (i-2) to (h-3);
		\draw[->] (i-3) to (h-1);
		\draw[->] (i-3) to (o-2);
		\draw[->] (i-4) to (h-1);
		\draw[->] (i-4) to (h-3);
		\draw[->] (h-1) to (o-1);
		\draw[->] (h-2) to (o-1);
		\draw[->] (h-2) to (o-2);
		\draw[->] (h-3) to (o-2);
		\draw[->] (o-2) edge[bend left] (h-3);

		\end{tikzpicture}
	}
	\hfill
	\chromosome{Outward}{0.2\columnwidth}
	{
		\begin{tabular}{r|l}
			1 & 6, 7, 8 \\
			2 & 7, 9 \\
			3 & 6, 7 \\
			4 & 7, 9 \\
			5 &  \\
			6 &  \\
			7 & 5 \\
			8 & 5, 6 \\
			9 & 6 \\
		\end{tabular}
	}
	\hfill
	\chromosome{Inward}{0.2\columnwidth}
	{
		\begin{tabular}{r|l}
			1 & \\
			2 & \\
			3 & \\
			4 & \\
			5 & 7, 8 \\
			6 & 1, 3, 8, 9 \\
			7 & 1, 2, 3, 4 \\
			8 & 1 \\
			9 & 2, 4 \\
		\end{tabular}
	}
	\hfill
	\chromosome{Recurrent}{0.2\columnwidth}
	{
		\begin{tabular}{r|l}
			1 & \\
			2 & \\
			3 & \\
			4 & \\
			5 & \\
			6 & 9 \\
			7 & \\
			8 & \\
			9 & \\
		\end{tabular}
	}
	\caption{(a) A phenotype with an unstructured topology. (b)--(d) Genotype of the phenotype in (a). The encoding is direct and without any redundancies.
		 Inward and recurrent connections are stored separately from outward connections to facilitate the evaluation of the phenotype (Fig.~\ref{fig:cortex_evaluation}).}
	\label{fig:cortex_genotype}
\end{figure}
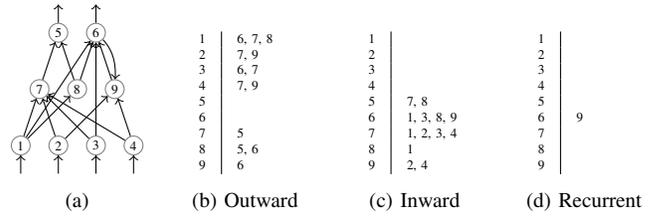

\begin{figure}
	\centering
	\subcaptionbox{Parent 1}[0.3\columnwidth]
	{
		\centering
		\begin{tikzpicture}[baseline,xscale=0.5,yscale=0.75] 

			\foreach \l [count=\i] in {1,2,3,4}
				\node [black_neuron] (bi-\i) at (\i,0) {\l};

			\foreach \l [count=\i] in {7,8,9}
				\node [black_neuron] (bh-\i) at (\i + 0.5,1) {\l};

			\foreach \l [count=\i] in {5,6}
				\node [black_neuron] (bo-\i) at (\i + 1,2) {\l};

			\draw[black!50] (bi-1) to (bh-1);
			\draw[black!50] (bi-1) to (bh-2);
			\draw[black!50] (bi-1) to (bo-2);
			\draw[black!50] (bi-2) to (bh-1);
			\draw[black!50] (bi-2) to (bh-3);
			\draw[black!50] (bi-3) to (bh-1);
			\draw[black!50] (bi-3) to (bo-2);
			\draw[black!50] (bi-4) to (bh-1);
			\draw[black!50] (bi-4) to (bh-3);
			\draw[black!50] (bh-1) to (bo-1);
			\draw[black!50] (bh-2) to (bo-1);
			\draw[black!50] (bh-2) to (bo-2);
			\draw[black!50] (bh-3) to (bo-2);
		\end{tikzpicture}

		\vspace{0.5em}
		{
			\tiny
			\centering
			\begin{tabular}[b]{r|l}
				1 & 6, 7, 8 \\
				2 & 7, 9 \\
				3 & 6, 7 \\
				4 & 7, 9 \\
				5 &  \\
				6 &  \\
				7 & 5 \\
				8 & 5, 6 \\
				9 & 6 \\
			\end{tabular}
		}
	}
	\subcaptionbox{Parent 2}[0.3\columnwidth]
	{
		\centering
		\begin{tikzpicture}[xscale=0.5,yscale=0.75] 

			\foreach \l [count=\i] in {1,2,3,4}
				\node [black_neuron] (ri-\i) at (\i,0) {\l};

			\foreach \l [count=\i] in {7,8,9}
				\node [black_neuron] (rh-\i) at (\i + 0.5,1) {\l};

			\foreach \l [count=\i] in {5,6}
				\node [black_neuron] (ro-\i) at (\i + 1,2) {\l};

			\draw[red_link] (ri-1) to (rh-1);
			\draw[red_link] (ri-1) to (rh-3);
			\draw[red_link] (ri-2) to (rh-2);
			\draw[red_link] (ri-2) to (rh-3);
			\draw[red_link] (ri-3) to (rh-1);
			\draw[red_link] (ri-3) to (rh-2);
			\draw[red_link] (ri-4) to (rh-1);
			\draw[red_link] (ri-4) to (rh-3);
			\draw[red_link] (rh-1) to (ro-1);
			\draw[red_link] (rh-2) to (ro-2);
			\draw[red_link] (rh-3) to (ro-2);

		\end{tikzpicture}

		\vspace{0.5em}
		{
			\tiny
			\centering
			\begin{tabular}[b]{r|l}
				\red{1} & \red{7}, \red{9}\\
				\red{2} & \red{8}, \red{9}\\
				\red{3} & \red{7}, \red{8}\\
				\red{4} & \red{7}, \red{9}\\
				\red{5} & \\
				\red{6} & \\
				\red{7} & \red{5} \\
				\red{8} & \red{6} \\
				\red{9} & \red{6} \\
			\end{tabular}
		}
	}
	\subcaptionbox{Offspring}[0.3\columnwidth]
	{
		\centering
		\begin{tikzpicture}[xscale=0.5,yscale=0.75] 

			\foreach \l [count=\i] in {1,2,3,4}
				\node [black_neuron] (off_i-\i) at (\i,0) {\l};

			\foreach \l [count=\i] in {7,8,9}
				\node [black_neuron] (off_h-\i) at (\i + 0.5,1) {\l};

			\foreach \l [count=\i] in {5,6}
				\node [black_neuron] (off_o-\i) at (\i + 1,2) {\l};

			\draw[black!50] (off_i-1) to (off_o-2);
			\draw[black!50] (off_i-1) to (off_h-1);
			\draw[red_link] (off_i-1) to (off_h-2);
			\draw[red_link] (off_i-1) to (off_h-3);
			\draw[black!50] (off_i-2) to (off_h-1);
			\draw[red_link] (off_i-2) to (off_h-2);
			\draw[black!50] (off_i-2) to (off_h-3);
			\draw[black!50] (off_i-3) to (off_o-2);
			\draw[black!50] (off_i-3) to (off_h-1);
			\draw[red_link] (off_i-3) to (off_h-2);
			\draw[red_link] (off_i-4) to (off_h-1);
			\draw[black!50] (off_i-4) to (off_h-3);
			\draw[black!50] (off_h-1) to (off_o-1);
			\draw[black!50] (off_h-2) to (off_o-1);
			\draw[red_link] (off_h-2) to (off_o-2);
			\draw[red_link] (off_h-3) to (off_o-2);
		\end{tikzpicture}

		\vspace{0.5em}
		{
			\tiny
			\centering
			\begin{tabular}[b]{r|l}
				1 & 6, 7, \red{8}, \red{9} \\
				2 & 7, \red{8}, 9 \\
				3 & 6, 7, \red{8} \\
				4 & \red{7}, 9 \\
				5 &  \\
				6 &  \\
				7 & 5 \\
				8 & 5, \red{6} \\
				9 & \red{6} \\
			\end{tabular}
		}
	}
	\caption{Crossover operation in Cortex. (a,b) Parents participating in the crossover operation and (c) the resulting offspring.
	Although only the outward connectivity table is shown here, crossover for recurrent connections is performed in the same manner.
	The chromosome of inward connections is populated automatically from the one of outward connections and does not participate in crossover.}
	\label{fig:cortex_crossover}
\end{figure}
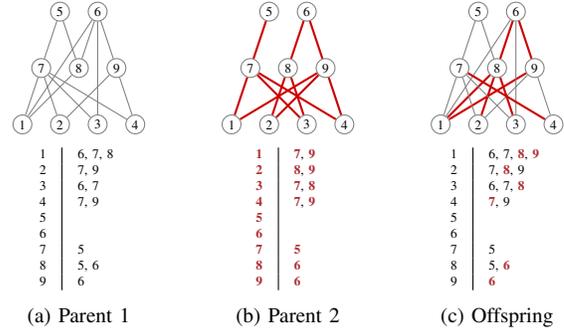

\subsubsection{Mutation}
The genotype representation requires that all neurons be numbered sequentially in the order of bias, input, output, hidden.
This allows for straightforward addition and deletion of connections by simply populating the inward and outward tables when adding a feedforward connection
or the recurrent table when adding a recurrent connection. It should be noted that before adding a feedforward connection, a check is performed to ensure that it would not form a loop.

Adding a hidden neuron is also straightforward.
Similarly to NEAT, the new neuron is appended to the end of the genotype and given the next neuron ID (Fig.~\ref{fig:cortex_add_neuron}), after which it is connected randomly
to an input and an output neuron. This avoids the overhead of having to check for accidental loops introduced by the new connections.

Deleting a neuron is somewhat more involved (Fig.~\ref{fig:cortex_delete_neuron}).
First, a hidden neuron is selected at random, and its connections are redistributed by connecting each of its inputs to each of its
outputs while ensuring that this does not introduce loops in the graph.
Subsequently, the neuron is deleted, and the IDs of all neurons with IDs greater than the deleted one are shifted down by 1 in order to maintain their consecutive numbering.
Finally, the same procedure of shifting IDs down is performed for all outward, inward and recurrent connections.

\begin{figure}
	\centering
	\subcaptionbox{Phenotype}[0.25\columnwidth]
	{
		\centering
		\begin{tikzpicture}[baseline,xscale=0.5,yscale=0.75]

			\foreach \l [count=\i] in {1,2,3,4}
				\node [black_neuron] (i-\i) at (\i,0) {\l};

			\foreach \l [count=\i] in {7,8,9}
				\node [black_neuron] (h-\i) at (\i + 0.5,1) {\l};

			\foreach \l [count=\i] in {5,6}
				\node [black_neuron] (o-\i) at (\i + 1,2) {\l};

			\node [red_neuron] (new) at (4.25,1) {10};

			\draw[black!50] (i-1) to (h-1);
			\draw[black!50] (i-1) to (h-2);
			\draw[black!50] (i-1) to (o-2);
			\draw[black!50] (i-2) to (h-1);
			\draw[black!50] (i-2) to (h-3);
			\draw[black!50] (i-3) to (h-1);
			\draw[black!50] (i-3) to (o-2);
			\draw[black!50] (i-4) to (h-1);
			\draw[black!50] (i-4) to (h-3);
			\draw[black!50] (h-1) to (o-1);
			\draw[black!50] (h-2) to (o-1);
			\draw[black!50] (h-2) to (o-2);
			\draw[black!50] (h-3) to (o-2);

			\draw[red_link] (i-3) -- (new);
			\draw[red_link] (new) -- (o-1);

		\end{tikzpicture}
	}
	\chromosome{Outward}{0.2\columnwidth}
	{
		\begin{tabular}[b]{r|l}
			1 & 6, 7, 8 \\
			2 & 7, 9 \\
			3 & 6, 7, \red{10} \\
			4 & 7, 9 \\
			5 &  \\
			6 &  \\
			7 & 5 \\
			8 & 5, 6 \\
			9 & 6 \\
			\red{10} & \red{5} \\
		\end{tabular}
	}
	\chromosome{Inward}{0.2\columnwidth}
	{
		\begin{tabular}[b]{r|l}
			1 & \\
			2 & \\
			3 & \\
			4 & \\
			5 & 7, 8, \red{10} \\
			6 & 1, 3, 8, 9 \\
			7 & 1, 2, 3, 4 \\
			8 & 1 \\
			9 & 2, 4 \\
			\red{10} & \red{3} \\
		\end{tabular}
	}
	\chromosome{Recurrent}{0.2\columnwidth}
	{
		\begin{tabular}[b]{r|l}
			1 & \\
			2 & \\
			3 & \\
			4 & \\
			5 & \\
			6 & 9 \\
			7 & \\
			8 & \\
			9 & \\
			\red{10} & \\
		\end{tabular}
	}
	\caption{(a) Adding a new hidden neuron to the phenotype. (b)--(d) The mutation is reflected in the genotype.
			The phenotype is placed in the species with four hidden neurons, which is created if it does not exist.
			Note that it is impossible to end up with a recurrent connection by accident.}
	\label{fig:cortex_add_neuron}
\end{figure}
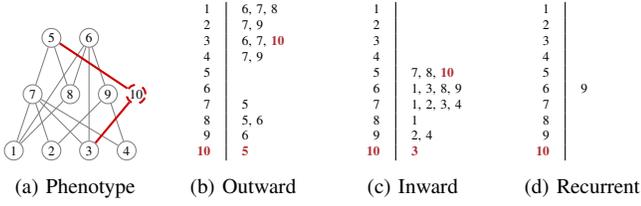

\begin{figure}
	\centering
	\subcaptionbox{}[0.23\columnwidth]
	{
		\centering
		\begin{tikzpicture}[xscale=0.5,yscale=0.75] 

			\foreach \l [count=\i] in {1,2,3,4}
				\node [black_neuron] (i-\i) at (\i,0) {\l};

			\node [black_neuron] (h-1) at (1.5,1) {7};
			\node [red_neuron] (h-2) at (2.5,1) {8};
			\node [black_neuron] (h-3) at (3.5,1) {9};

			\foreach \l [count=\i] in {5,6}
				\node [black_neuron] (o-\i) at (\i + 1,2) {\l};

			\draw[black!50] (i-1) to (h-1);
			\draw[red_link] (i-1) to (h-2);
			\draw[black!50] (i-1) to (o-2);
			\draw[black!50] (i-2) to (h-1);
			\draw[black!50] (i-2) to (h-3);
			\draw[black!50] (i-3) to (h-1);
			\draw[black!50] (i-3) to (o-2);
			\draw[black!50] (i-4) to (h-1);
			\draw[black!50] (i-4) to (h-3);
			\draw[black!50] (h-1) to (o-1);
			\draw[red_link] (h-2) to (o-1);
			\draw[red_link] (h-2) to (o-2);
			\draw[black!50] (h-3) to (o-2);
		\end{tikzpicture}

		\vspace{0.5em}
		{
			\tiny
			\centering
			\begin{tabular}[b]{r|l}
				1 & 6, 7, 8 \\
				2 & 7, 9 \\
				3 & 6, 7 \\
				4 & 7, 9 \\
				5 &  \\
				6 &  \\
				7 & 5 \\
				\red{\textbf{8}} & \red{\textbf{5}}, \red{\textbf{6}} \\
				9 & 6 \\
			\end{tabular}
		}
	}
	\subcaptionbox{}[0.23\columnwidth]
	{
		\centering
		\begin{tikzpicture}[xscale=0.5,yscale=0.75] 

			\foreach \l [count=\i] in {1,2,3,4}
				\node [black_neuron] (i-\i) at (\i,0) {\l};

			\node [black_neuron] (h-1) at (1.5,1) {7};
			\node [red_neuron] (h-2) at (2.5,1) {8};
			\node [black_neuron] (h-3) at (3.5,1) {9};

			\foreach \l [count=\i] in {5,6}
				\node [black_neuron] (o-\i) at (\i + 1,2) {\l};

			\draw[black!50] (i-1) to (h-1);
			\draw[black!50] (i-1) to (o-2);
			\draw[black!50] (i-2) to (h-1);
			\draw[black!50] (i-2) to (h-3);
			\draw[black!50] (i-3) to (h-1);
			\draw[black!50] (i-3) to (o-2);
			\draw[black!50] (i-4) to (h-1);
			\draw[black!50] (i-4) to (h-3);
			\draw[black!50] (h-1) to (o-1);
			\draw[black!50] (h-3) to (o-2);
			\draw[red_link] (i-1) edge[bend left] (o-1);
		\end{tikzpicture}

		\vspace{0.5em}
		{
			\tiny
			\centering
			\begin{tabular}[b]{r|l}
				1 & \red{5}, 6, 7 \\
				2 & 7, 9 \\
				3 & 6, 7 \\
				4 & 7, 9 \\
				5 &  \\
				6 &  \\
				7 & 5 \\
				\red{8} & \\
				9 & 6 \\
			\end{tabular}
		}
	}
	\subcaptionbox{}[0.25\columnwidth]
	{
		\centering
		\begin{tikzpicture}[xscale=0.5,yscale=0.75] 

			\foreach \l [count=\i] in {1,2,3,4}
			\node [black_neuron] (i-\i) at (\i,0) {\l};

			\node [black_neuron] (h-1) at (1.5,1) {7};
			\node [red_neuron] (h-3) at (3.5,1) {9};

			\foreach \l [count=\i] in {5,6}
			\node [black_neuron] (o-\i) at (\i + 1,2) {\l};

			\draw[black!50] (i-1) to (h-1);
			\draw[black!50] (i-1) to (o-2);
			\draw[black!50] (i-2) to (h-1);
			\draw[black!50] (i-2) to (h-3);
			\draw[black!50] (i-3) to (h-1);
			\draw[black!50] (i-3) to (o-2);
			\draw[black!50] (i-4) to (h-1);
			\draw[black!50] (i-4) to (h-3);
			\draw[black!50] (h-1) to (o-1);
			\draw[black!50] (h-3) to (o-2);
			\draw[black!50] (i-1) edge[bend left] (o-1);
		\end{tikzpicture}

		\vspace{0.5em}
		{
			\tiny
			\centering
			\begin{tabular}[b]{r|l}
				1 & 5, 6, 7 \\
				2 & 7, 9 $\mapsto$ \red{8} \\
				3 & 6, 7 \\
				4 & 7, 9 $\mapsto$ \red{8} \\
				5 &  \\
				6 &  \\
				7 & 5 \\
				9 $\mapsto$ \red{8} & 6 \\
				\multicolumn{2}{c}{} \\
			\end{tabular}
		}
	}
	\subcaptionbox{}[0.23\columnwidth]
	{
		\centering
		\begin{tikzpicture}[xscale=0.5,yscale=0.75] 

			\foreach \l [count=\i] in {1,2,3,4}
			\node [black_neuron] (i-\i) at (\i,0) {\l};

			\node [black_neuron] (h-1) at (1.5,1) {7};
			\node [black_neuron] (h-3) at (3.5,1) {8};

			\foreach \l [count=\i] in {5,6}
			\node [black_neuron] (o-\i) at (\i + 1,2) {\l};

			\draw[black!50] (i-1) to (h-1);
			\draw[black!50] (i-1) to (o-2);
			\draw[black!50] (i-2) to (h-1);
			\draw[black!50] (i-2) to (h-3);
			\draw[black!50] (i-3) to (h-1);
			\draw[black!50] (i-3) to (o-2);
			\draw[black!50] (i-4) to (h-1);
			\draw[black!50] (i-4) to (h-3);
			\draw[black!50] (h-1) to (o-1);
			\draw[black!50] (h-3) to (o-2);
			\draw[black!50] (i-1) edge[bend left] (o-1);
		\end{tikzpicture}

		\vspace{0.5em}
		{
			\tiny
			\centering
			\begin{tabular}[b]{r|l}
				1 & 5, 6, 7 \\
				2 & 7, 8 \\
				3 & 6, 7 \\
				4 & 7, 8 \\
				5 &  \\
				6 &  \\
				7 & 5 \\
				8 & 6 \\
				\multicolumn{2}{c}{} \\
			\end{tabular}
		}
	}
	\caption{Deleting a neuron from a phenotype. Only the chromosome of outward connections is shown, but the same procedure is applied to the other two chromosomes.
		(a) First, a hidden neuron is selected at random. (b) All outward connections from the selected neuron are
		transferred to each source neuron if they do not exist already.
		At this stage, if a transferred connection would introduce a loop (a recurrent connection), it is not expressed.
		(c) Now the selected neuron does not have any connections and can be safely removed. However, this leaves the
		genotype in an inconsistent state since the neuron numbering must be consecutive.
		Therefore, all neurons whose ID is higher than the deleted neuron, as well as all connections to and from them, are shifted down by 1.
		(d) The new phenotype now has a consistent genotype and belongs to a different species.}
	\label{fig:cortex_delete_neuron}
\end{figure}
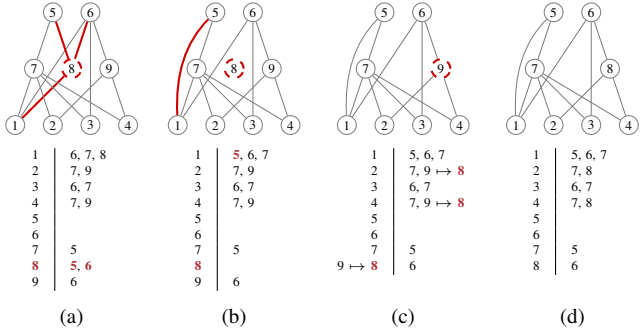

It should be noted that adding or deleting a neuron results in the network being moved to the species with the corresponding number of hidden neurons.
If such a species does not exist, it is created.
In this way, search spaces of incrementally larger size are being explored \textit{simultaneously} by different populations of networks,
which compete with each other indirectly at the species level.
The purpose of this scheme is to converge on the optimal number of hidden neurons necessary for performing the task at hand.

Weights are mutated by selecting a random connection and changing its weight according to the following procedure.
The first mutation for a given weight $w$ is in a random direction (increase or decrease), where the new weight value is drawn from a
normal distribution with a mean at the current value and a relatively large standard deviation $\sigma_{w}$ (equivalent to about 5\% of the allowable weight range).
The rationale behind using a large initial value for $\sigma_{w}$ is that in the early stages individuals are much more likely
to be in a low-fitness part of the search space than close to an optimum, and therefore exploration should initially dominate exploitation.
However, only the first mutation for each weight is in a random direction.
After each mutation, the fitness of the individual is re-evaluated, and the difference from the old fitness is noted.
If the fitness has improved, this means that the mutation has been beneficial, and the direction of weight update is maintained.
Conversely, if the fitness has decreased, the mutation must have been unproductive,
and the direction is reversed while at the same time reducing the magnitude of $\sigma_{w}$ by a small fraction.
This is a form of \textit{meta-optimisation} which is a hybrid of two existing techniques for heuristic optimisation:
pattern search (PS) and the Luus-Jaacola (LJ) method \parencite{Luus_Jaakola--1973,Pedersen--2010}.
In the PS method, the direction of weight update is maintained until the fitness stops increasing,
at which point the direction is reversed and the sampling range is halved.
The LJ method is an extension of the PS method in which \textit{all} weights are updated simultaneously
using a common sampling range (the standard deviation $\sigma_{w}$ if the values
are drawn from a normal distribution), which is decreased by a small fraction every time the fitness fails to improve.

In Cortex, the PS and LJ methods are combined as follows.
For a given network, only a single (randomly selected) weight is updated in each generation, and the direction of change is maintained
in subsequent mutations of the same weight until the fitness fails to improve.
However, instead of being halved at this point, $\sigma_{w}$ is decreased gradually by being multiplied by a fixed coefficient $0 < d < 1$
(currently, the default value of $d$ is $0.95$, in accordance with \parencite{Luus_Jaakola--1973}).
In this way, it is possible to determine exactly what mutations are beneficial, without compromising the initial tendency towards exploration versus exploitation.
With the original PS method, if the very first weight update is \textit{not} beneficial, the sampling range would be instantly halved,
potentially reducing the speed of convergence. This problem is largely avoided if the sampling range is decreased at a slower exponential rate.

\subsubsection{Stagnancy}

If the fitness of an individual fails to improve for a certain number of generations (denoted as $g_{max}$ in Eq.~\ref{eq:stagnancy}), its fitness is gradually reduced
by the stagnancy coefficient $C_{d}$ as follows:

\begin{equation}\label{eq:stagnancy}
C_{d} = \begin{cases}
				1, & g \leq g_{max}\\
				1 - \exp\left({ -\left(1 + \frac{1}{g}\right)^{2}}\right), & g > g_{max}
			\end{cases},
\end{equation}

\begin{figure}
	\centering
	\includegraphics[width=\columnwidth]{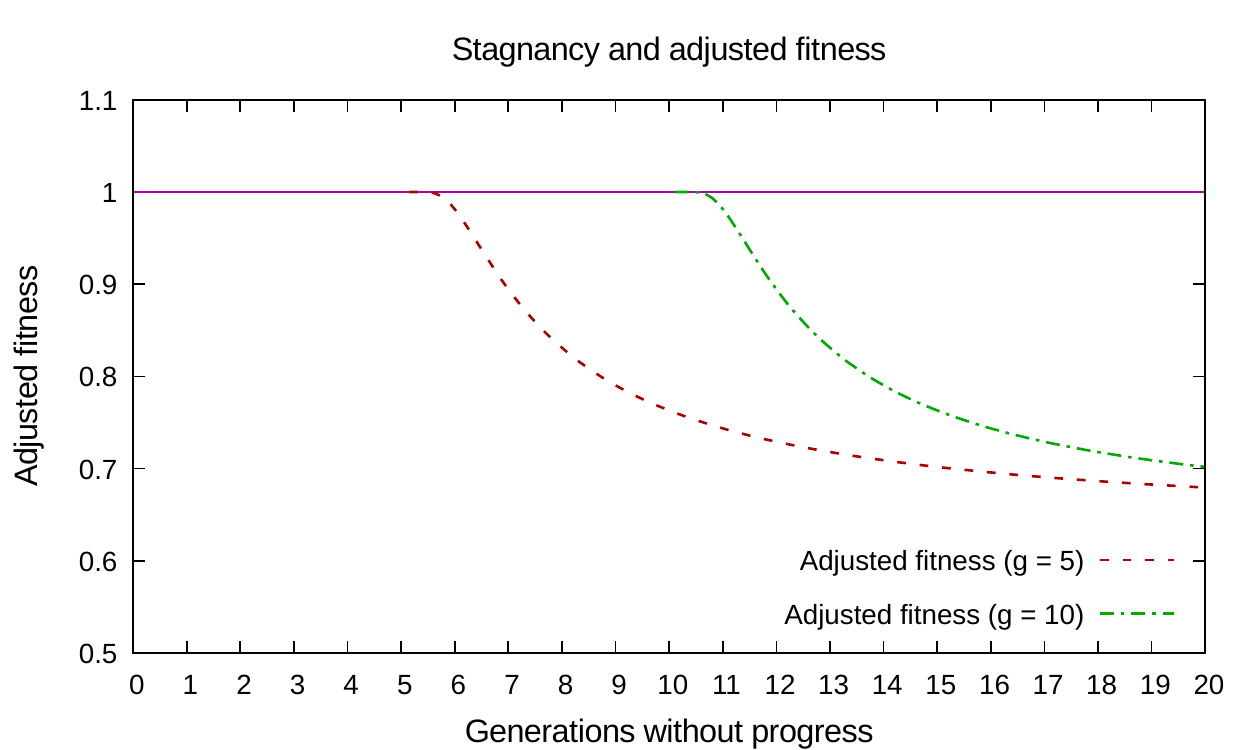}
	\caption{The adjusted fitness of an individual is gradually decreased if it fails to make progress after a certain number of generations.}
	\label{fig:adjusted_fitness}
\end{figure}

Here, $g$ is the number of generations since the last fitness improvement.
Essentially, the network is gradually penalised if it fails to make any progress after several generations (a stagnancy-based penalty scheme is also employed in NEAT).
Using $C_{d}$, we calculate the adjusted fitness $\tilde{f_{i}}$ of each individual
by normalising the true fitness $f_{i}$ to the highest fitness $f_{max}$ achieved in the entire ecosystem (Fig.~\ref{fig:adjusted_fitness}):

\begin{equation}\label{eq:adjusted_fitness}
	\tilde{f_{i}} = \frac{C_{d} f_{i}}{f_{max}}.
\end{equation}

In the operations involved in offspring generation, we utilise the adjusted fitness to rank individuals.

\subsubsection{Offspring Generation}
The crossover operator is applied separately to each species rather than to the ecosystem as a whole.
In addition to individual fitness, each species is also assigned a fitness value based on the average fitness of the individuals belonging to it:

\begin{equation}
	f_{s_{i}} = \frac{1}{n}\sum_{j=1}^{n}{f_{j}},
\end{equation}

where $f_{s_{i}}$ is the fitness of species $i$, $f_{j}$ is the fitness of the $j$-th individual of the same species,
and $n$ is the number of individuals belonging to that species.
The adjusted fitness $\tilde{f_{s_{i}}}$ of species $i$ is defined as the sum of the
adjusted fitness values of all individuals belonging to that species:

\begin{equation}
	\tilde{f_{s_{i}}} = \sum_{j}{}{\tilde{f_{j}}}
\end{equation}

Similarly to the case of individuals, species are also ranked according to their adjusted fitness.
Another metric which is recalculated at each generation for each species is \textit{diversity}.
The diversity $d_{i}$ of species $i$ represents the variance of the fitness values of all individuals belonging to that species:

\begin{equation}
	d_{i} = \frac{1}{n}\sum_{j=1}^{n}{\left(\frac{f_{j} - f_{s_{i}}}{f_{max} - f_{min}}\right)^{2}},
\end{equation}
where $f_{max}$ and $f_{min}$ are the maximal and minimal individual fitness values for species $i$, and $n$ is the number of individuals (population size) for that species.

At each generation, the adjusted species fitness $\tilde{f_{s_{i}}}$ is employed to determine the number of offspring that each species is allowed produce,
and the diversity is used to determine how many individuals from the species are allowed to reproduce, in other words, the number of parents.
In this way, Cortex employs a double screening to ensure diversity. For example, on one hand, very fit species are given a large quota for offspring,
but if the diversity within that species is low, the number of parents will also be low, and the actual number of offspring produced is unlikely to even reach the quota.
Conversely, a species with low fitness but large diversity will be assigned a large parenting quota but a small offspring quota, and thus the resulting total number of
offspring will be likewise small. In order to produce a large number of offspring, a species must be very fit \textit{as well as} very diverse.

Note that the offspring quota $q$ serves only as an upper limit and does not mean that $q$ individuals are necessarily removed at each generation.
Rather, $q$ is distributed among the species according to the following formula:

\begin{equation}
	C_{q} = \frac{q}{\sum_{i=1}^{n}{\left(1 + \ln{\tilde{f_{s_{i}}} } \right)}}
\end{equation}

\begin{equation}
	o_{i} = \left\lfloor C_{q} \ln{\tilde{f_{s_{i}}}} \right\rfloor
\end{equation}

Here, $\tilde{f_{s_{i}}}$ is the adjusted species fitness as defined above, $C_{q}$ is a quota distribution coefficient common to all species, and $o_{i}$ is the number of
offspring that species $i$ is allowed to produce.
This distribution scheme promotes diversity by ensuring that species with similar fitness
will be given similar opportunities to reproduce while reducing the risk of a single species taking over the entire ecosystem.
The parenting quota is determined in a similar manner, but using diversity instead of the adjusted fitness.

The ecosystem size in Cortex is fixed in order to prevent uncontrolled expansion,
and the total number of individuals from all species cannot exceed the predefined limit.
The global offspring and parenting quotas are taken as percentages of the maximal ecosystem size.
Note that this percentage is the same for both the offspring and parenting quotas.
At each generation, each species produces a number of offspring depending on its diversity and fitness, and if the total number of individuals
and offspring exceeds the maximal allowable size of the ecosystem, an elimination procedure is employed to bring the ecosystem size to within the limit.
The elimination proceeds by selecting a random species according to the roulette wheel principle and removing the individual with the lowest fitness from it.
This is repeated until the total number of individuals is equal to the maximal ecosystem size.
The weight $w_{i}$ for species $i$ in the roulette wheel (a weighted distribution) is calculated as follows:

\begin{equation}
	w_{i} = \frac{s_{i}p_{i}}{d_{i}},
\end{equation}

where $s_{i}$ is the stagnancy of the species (the number of generations for which none of the individuals in that species made any progress in terms of fitness),
$p_{i}$ is the population size of the species, and $d_{i}$ is the diversity of the species as defined above.
Again, this metric is designed to promote diversity by increasing the probability of elimination for individuals from stagnant, overly populous and less diverse species
while being gentle on progressive, small and diverse species. However, the stochastic nature of the roulette wheel also ensures that no species is immune to total extermination.

\subsection{Network Evaluation}
\begin{figure}
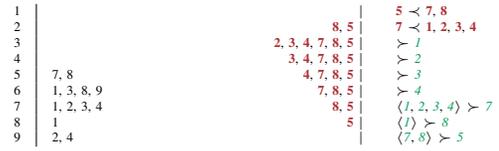

	\centering
	\chromosome{Inward connections}{0.3\columnwidth}
	{
		\centering
		\begin{tabular}{l|l}
			1 & \\
			2 & \\
			3 & \\
			4 & \\
			5 & 7, 8 \\
			6 & 1, 3, 8, 9 \\
			7 & 1, 2, 3, 4 \\
			8 & 1 \\
			9 & 2, 4 \\
		\end{tabular}
	}
	\subcaptionbox{Evaluation stack for output neuron 5.}[0.6\columnwidth]
	{
		\tiny
		\centering
		\begin{tabular}{rl}
				$|$ & \red{5} $\prec$ \red{7}, \red{8} \\
				\red{8}, \red{5} $|$ & \red{7} $\prec$ \red{1}, \red{2}, \red{3}, \red{4} \\
				\red{2}, \red{3}, \red{4}, \red{7}, \red{8}, \red{5} $|$ & $\succ$ \green{1} \\
				\red{3}, \red{4}, \red{7}, \red{8}, \red{5} $|$ & $\succ$ \green{2} \\
				\red{4}, \red{7}, \red{8}, \red{5} $|$ & $\succ$ \green{3} \\
				\red{7}, \red{8}, \red{5} $|$ & $\succ$ \green{4} \\
				\red{8}, \red{5} $|$ & $\left\langle \text{\green{1}, \green{2}, \green{3}, \green{4}} \right\rangle$ $\succ$ \green{7} \\
				\red{5} $|$ & $\left\langle \text{\green{1}} \right\rangle$ $\succ$ \green{8} \\
				$|$ & $\left\langle \text{\green{7}, \green{8}} \right\rangle$ $\succ$ \green{5} \\
		\end{tabular}
	}
	\caption{Stack-based phenotype evaluation procedure. (a) Inward connections for the network in Fig.~\ref{fig:cortex_genotype}. (b) Evaluation stack for output neuron 5 in (a). It is straightforward to derive the order in which neurons have to be evaluated before producing the final output. The evaluation proceeds backwards from the output neurons.
	If there are prerequisites (source neurons) which have not been evaluated, they are pushed onto the stack, and \textit{their} respective prerequisites are evaluated.
	This is performed recursively until all input neurons are evaluated (input neurons have no prerequisites,
	indicated by the corresponding empty fields in the chromosome of inward connections).
	This procedure allows for efficient and reproducible evaluation of networks with arbitrary topology.
	After all output values are computed, activations are propagated along any recurrent connections in preparation for the next evaluation step.}
	\label{fig:cortex_evaluation}
\end{figure}

As mentioned above, the genotype is used directly in the evaluation of the phenotype.
Due to the ordered nature of the genotype, it is clear which neurons must be evaluated before an output is produced.
The evaluation procedure is outlined in Fig.~\ref{fig:cortex_evaluation}.

Although the network evaluation procedure is rarely mentioned explicitly in NE research, it is noteworthy that the evaluation process employed in some NE
frameworks is incremental, meaning that the network becomes activated gradually over several time steps.
However, this poses the problem that the network may never actually become stable \parencite{Montana_Wyk_Brinn_Montana_Milligan--2009}.
For example, for many applications, NEAT requires that the network `settle' on a value (or even a value within some small delta \parencite{Staney--NEAT_homepage})
before it is considered to have produced an output.
Clearly, this evaluation scheme does not always provide a reproducible output.
In other words, when presented with the same input more than once, the network may produce
different outputs, or it may never even settle on an output at all.
This issue is addressed by the stack-based network evaluation procedure in Cortex, which ensures that the output of the network is reproducible
even for networks with arbitrary topologies and recurrent connections.
Specifically, the entire network is evaluated at every step. Once all outputs are evaluated, signals travelling via recurrent connections
(if any) are propagated in preparation for the next evaluation step when the next external input arrives.
This is the reason for storing recurrent connections in a separate chromosome.

\section{Experiments}\label{sec:experiments}
The advantages of the speciation principle and the overall performance of Cortex were evaluated experimentally by using a
software platform written in C++ which implements the proposed framework\footnote{The latest version of Cortex is available at \url{https://github.com/trilobeat/cortex}}.

\subsection{XOR}
As discussed above, certain tasks require networks to search a space of sufficient complexity in order to be able to find a solution reliably.
In this regard, the proposed speciation principle allows the ecosystem to explore a range of search spaces of increasing complexity simultaneously.
Since the speciation criterion is known \textit{a priori}, the ecosystem can be initialised with more than one species,
allowing it to explore spaces of different complexity \textit{from the onset of evolution}.

A simple and clear way to test the effectiveness of this idea is the XOR task, which is not linearly separable and thus requires at least one hidden neuron.
Ten experiments (referred to as `evolving speciation' experiments below) were performed on the XOR task by using an ecosystem randomly initialised with
1--10 species (0--9 hidden neurons).
Ten additional experiments were performed with the addition and deletion of neurons disabled (`fixed speciation' experiments)
in order to test the impact of dynamic speciation on the speed of finding a solution and the success rate.
Each experiment consisted of 100 runs, and a run was terminated if no solution was found within 100 generations.
The ecosystem size limit was 150 networks, and the global offspring / parenting quota was 25\%
of this limit.
At the onset of each simulation, the 150 networks were evenly distributed among the available species.

\begin{figure}
\centering
	\subcaptionbox{\label{fig:xor_success_rate}}[\columnwidth]
	{
		\centering
		\includegraphics[width=\columnwidth]{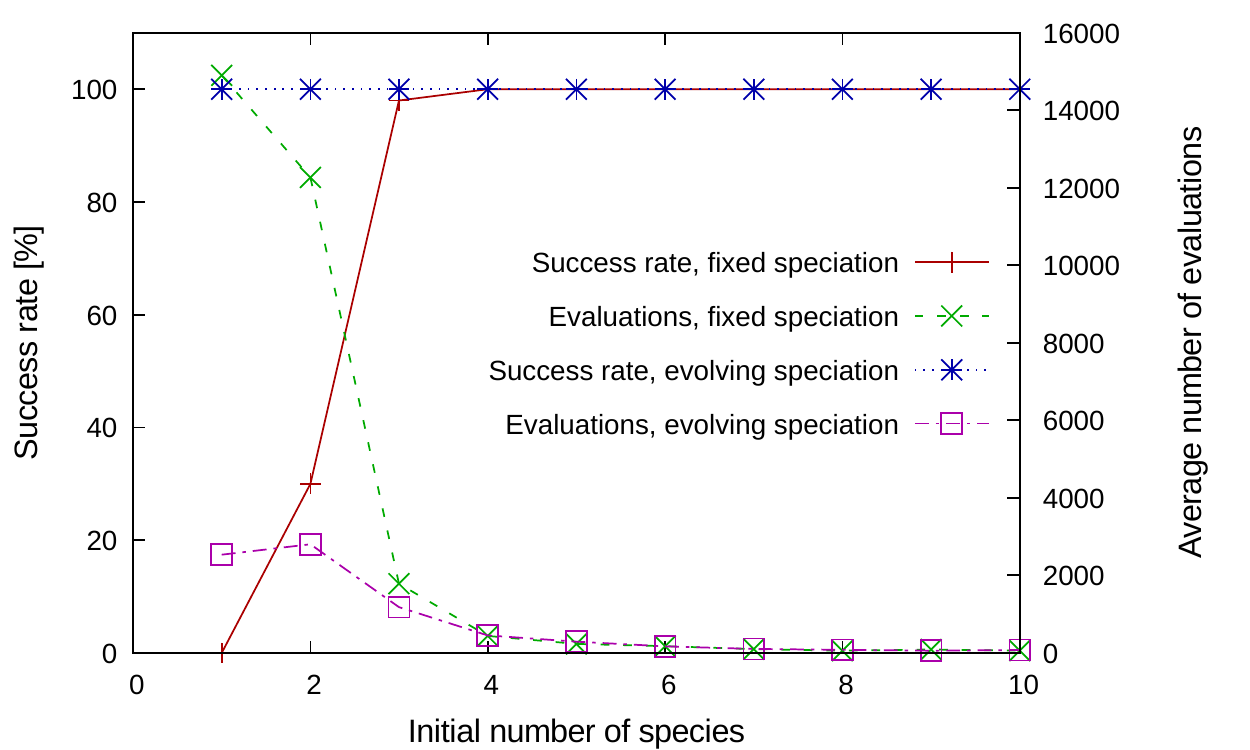}
	}
	\subcaptionbox{\label{fig:xor_average_hidden_neurons}}[\columnwidth]
	{
		\centering
		\includegraphics[width=\columnwidth]{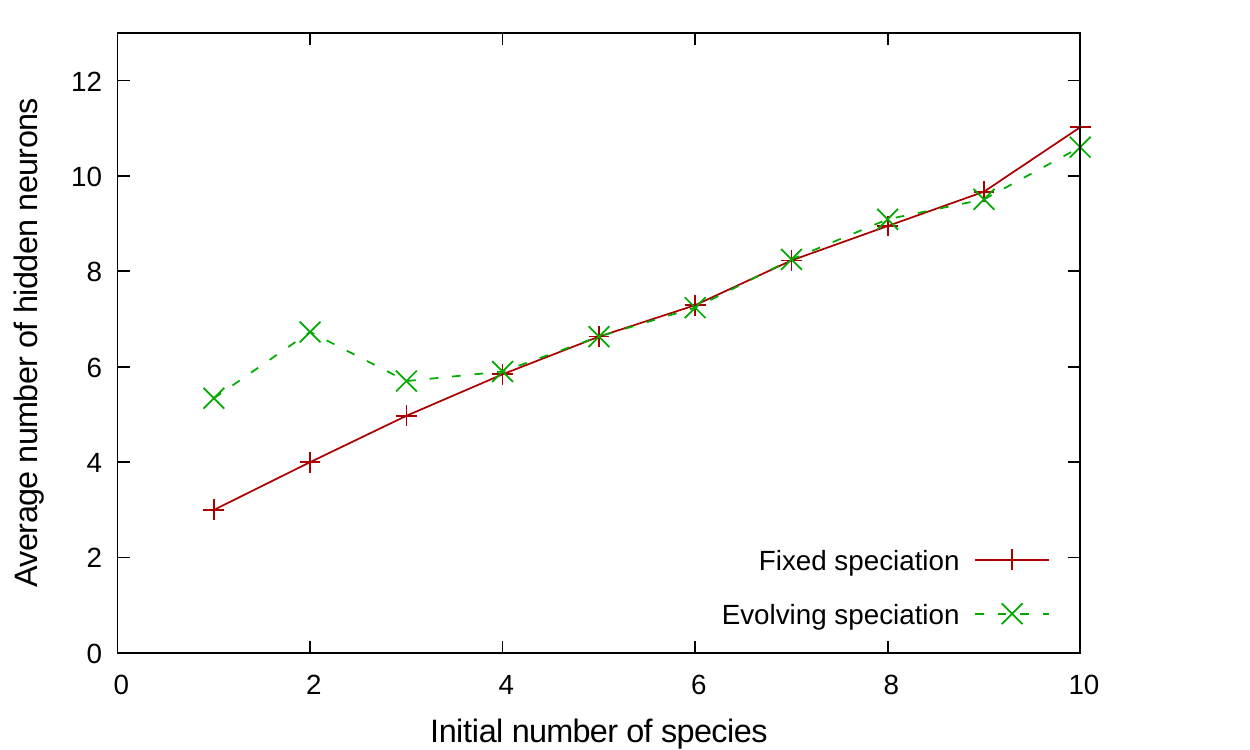}
	}
	\caption{(a) Results for the evolving and fixed speciation experiments on the XOR task.
		As expected, the ecosystem with one species (zero hidden neurons) in the fixed speciation experiments does not find a single solution in 100 runs.
		The situation is markedly improved with the addition of more species, and finding a solution is not a problem if there are four or more species.
		In contrast, a solution is always found in the evolving speciation experiments. Convergence is very rapid even if there is initially only a single species
		(average number of evaluations: 2535), but the same trend towards decreasing number of evaluations with increasing initial number of species is seen as in the fixed speciation experiments.
		(b) Average number of neurons in evolved solutions plotted against the number of species. The plots practically overlap when the ecosystem is initialised with four or more species,
		in tune with the data in (a).}
	\label{fig:xor}
\end{figure}

The probability weightings for adding and deleting a neuron
were 30 and 5 units, respectively, in the evolving speciation experiments, and 0 in the case of fixed speciation experiments.
Furthermore, the probability weighting of mutating a random weight was 1000 units,
and that for adding and deleting a connection was 50 and 5 units, respectively.
These parameters are virtually identical to those used in the XOR experiments using NEAT in \parencite{Stanley_Miikkulainen--2002}.
The elite size was 1, meaning that the champion network in each species was copied into the next generation unchanged.
At each generation, offspring was produced through crossover with a 75\% chance and through cloning combined with mutation with a 25\% chance,
and the remaining networks which were not part of the elite were mutated with the above parameters.
The $\tanh$ activation function was used for all hidden and output neurons,
whereby a positive output was interpreted as a $1$ and a non-negative output was interpreted as a $0$.
The fitness function was calculated as $4-\epsilon$, where $\epsilon$ is the sum of absolute errors for each of the four outputs.
The results are shown in Fig.~\ref{fig:xor}.

Clearly, the XOR task does not pose a significant problem if Cortex is allowed to speciate freely.
In the evolving speciation experiments, it finds a solution reliably in 100\% of the runs, even if it is initialised with a single species (no hidden neurons).
Looking at the success rate in Fig.~\ref{fig:xor_success_rate}, there is practically no difference in performance if the ecosystem
is initialised with four or more species with both fixed and evolving speciation.
Note that the average number of neurons for the two sets of experiments practically overlaps when the initial number of species is four or greater,
suggesting that the proposed speciation procedure can lead to a significant improvement in performance (constant 100\% success rate)
with virtually zero overhead in terms of the \textit{size} of the resulting solution.

One possible explanation for the low performance ($\sim30\%$ success rate) of Cortex with fixed speciation in the case of two species
($0$ and $1$ hidden neurons) is that the initial number of species determines the number of individuals standing a chance of solving the problem.
Thus, with two initial species, only 50\% of the individuals can solve the task, whereas with ten initial species 90\% of all individuals can potentially solve it.
In addition, any increase in the population of networks without hidden neurons further reduces the number of networks \textit{with} hidden neurons due to
the fixed size of the ecosystem.
This `dilution' illustrates the importance of diversity in terms of complexity, which can be ensured by allowing the ecosystem to spawn new species.

\subsection{Double Pole Balancing}
The XOR task was useful for illustrating the advantage of initialising an ecosystem with multiple species in cases where the solution depends on a minimal number of hidden neurons.
However, the XOR task is too simple to provide a good estimate of the overall performance of the proposed framework.
Therefore, experiments were conducted on the double pole balancing task with velocity information.
This task can be successfully solved by a neural network controller without any hidden neurons for a certain set of conditions,
and therefore it evaluates the performance of the proposed framework as a whole rather than
emphasising the advantages of speciation as in the preceding experiment.

The equations of motion of the cart and the two poles are given in \parencite{Wieland--1991}.
Experiments were conducted by taking into account all parameters, including friction coefficients.
The fourth-order Runge-Kutta method was used for accurate calculation of the trajectories.

\subsubsection{Fixed conditions} The first set of experiments were designed to verify whether Cortex could find a solution to the problem at all.
The conditions used in the experiments are shown in Table~\ref{table:dpb}. In the simulations, all variables presented as input to the controller
were scaled to the interval $[-1,1]$.

The first experiment used conditions which were similar to those reported in other studies
\parencite{Gruau_Whitley_Pyeatt--1996,Stanley_Miikkulainen--2002,Gomez_Miikkulainen--1999} in order to ensure that the comparison is performed on an equal footing.
Specifically, the initial angle of the long pole was taken at random from the interval $[-6$\textdegree,$6$\textdegree$]$, the initial velocity was $0\,m/s$,
and the simulation started with the cart located at the centre of the track ($2.4\,m$ from either end of the track).
The results are presented in Table~\ref{table:dpb_results} together with the results reported in \parencite{Stanley_Miikkulainen--2002} and \parencite{Gomez_Miikkulainen--1999}.
It is indicative that the number of hidden neurons was very low, showing that on average only one in four solutions required a hidden neuron.
The complexity of the solution is clearly low, in agreement with the conclusion in \parencite{Gruau_Whitley_Pyeatt--1996} that evolution can find
surprisingly simple solutions to problems considered very difficult by humans.
A human designing a controller for the double pole balancing task would therefore tend to overengineer the solution,
as exemplified by the use of a fully connected, fully recurrent network with $10$ hidden neurons in \parencite{Wieland--1991}.
Potentially, the large number of hidden neurons could make the network prone to overfitting, thus actually \textit{preventing} it from finding an optimal solution.

\begin{table}
	\caption{Conditions for double pole balancing experiments}
	\label{table:dpb}
	\centering
	\begin{tabular}{lrr}
		\hline \\
		& Unscaled & Scaled \\
		\hline \\
		Track length & $4.8\,m$ & \\
		Cart position relative to track centre & $\pm2.4\,m$ & $\pm1$ \\
		Initial velocity of the cart & $0\,m/s$ & $0$ \\
		Threshold angle & $\pm36$\textdegree & $\pm1$ \\
		Initial angle (long pole) & $\pm6$\textdegree & $\pm0.167$  \\
		Initial angle (short pole) & $0$\textdegree & $0$ \\
		Initial angular velocity (long and short poles) & $0\,rad/s$ & $0$ \\
		Time step (RK method parameter) & $0.01\,s$ & \\
		Maximal force magnitude & $10\,N$ & \\
		Coefficient of friction (cart/track) & $0.0005$ & \\
		Coefficient of friction (pole/hinge) & $0.000002$ & \\
		Cart mass & $1\,kg$ & \\
		Long pole mass & $0.1\,kg$ & \\
		Short pole mass & $0.01\,kg$ & \\
		Long pole length & $1.0\,m$ & \\
		Short pole length & $0.1\,m$ & \\
		\hline
	\end{tabular}
\end{table}

\begin{table}
	\caption{Results for double pole balancing with velocity}
	\label{table:dpb_results}
	\centering
	\begin{tabular}{lrrrr}
		\hline \\
		& Evaluations & Generations & Nets & Hidden neurons\\
		\hline \\
		SANE \parencite{Gomez_Miikkulainen--1999} & $12 600$ & 63 & 200 & -- \\
		ESP \parencite{Gomez_Miikkulainen--1999} & $3800$ & 19 & 200 & -- \\
		NEAT \parencite{Stanley_Miikkulainen--2002} & $3600$ & 24 & 150 & ? ($0$--$4$) \\
		Cortex & $2547 $ & 17 & 150 & 0.27 \\
		\hline

	\end{tabular}
\end{table}

\subsubsection{Generalisation Test}

After solving the task for the restricted set of conditions presented above, Cortex was subjected to a more rigorous \textit{generalisation} test to evaluate its performance
over a wider range of conditions. It should be noted that previous studies \parencite{Gruau_Whitley_Pyeatt--1996,Stanley_Miikkulainen--2002} have evaluated the
generalisation performance by testing all combinations of 0.05, 0.25, 0.5, 0.75 and 0.95 of the maximal values of the cart position, cart velocity,
long pole angle and long pole angular velocity, for a total of 625 (=5$^{4}$) experiments.

From a purely physical point of view, if the long pole is leaning at a large angle and the cart is located near the end of the track while travelling at high speed,
it is highly unlikely that the poles can be recovered to a stable region from which they could be balanced for an appreciable amount of time.
Performing such a test with deliberately impossible conditions was considered uninformative, and therefore a different approach to generalisation was adopted in this study.
The approach was similar to that reported in \parencite{Blair--2014}, but with several extensions.
First, a very simple experiment was designed with the aim to find the maximal angle from which the system could recover as a function of the initial position of the cart.
In this experiment, the initial velocity of the cart and the initial angular velocity of the long pole were set to $0$, and the maximal allowable force was applied in the same direction
until the cart reached the end of the track. The initial position of the cart was varied from $0$ to $2.4\,m$ in increments of $0.12\,m$, and the initial angle of the long pole
was varied from $1$\textdegree~to $36$\textdegree~in increments of $1$\textdegree.
Intuitively, the maximal angle from which the system can recover should gradually \textit{decrease} as the initial position approaches the end of the track.

\begin{figure}
	\centering
	\includegraphics[width=\columnwidth]{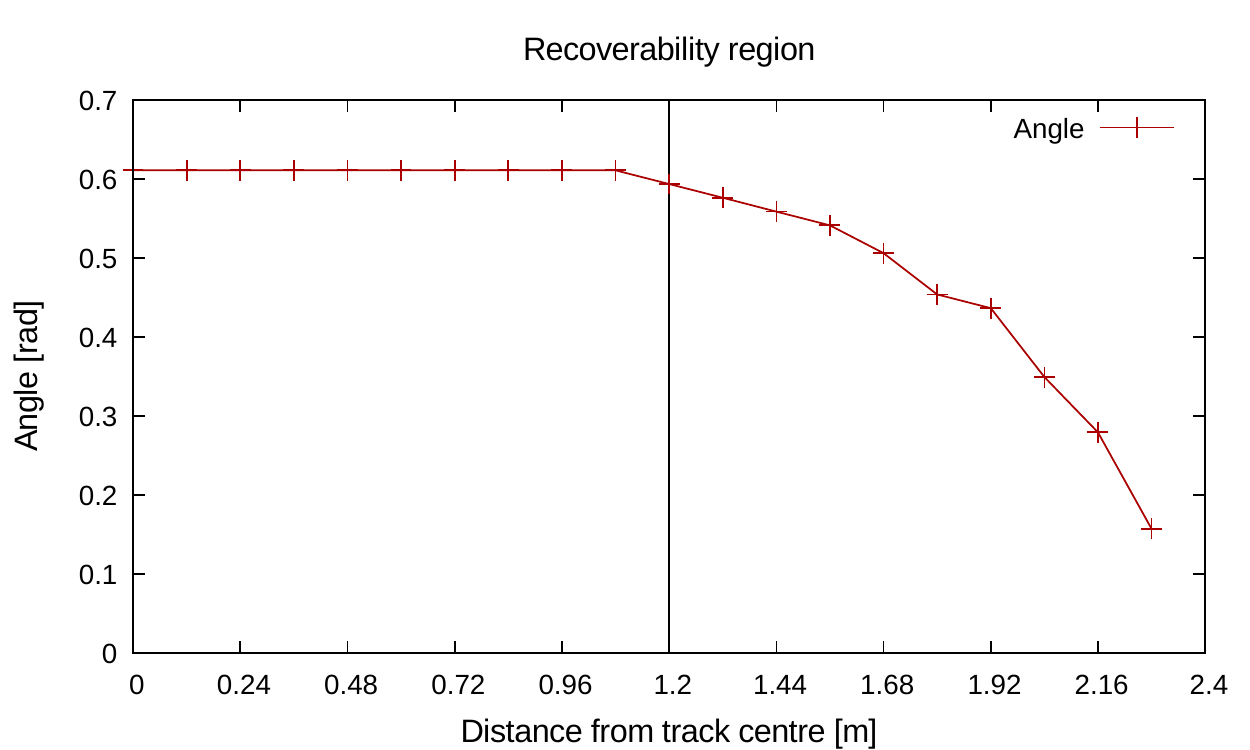}

	\caption{Maximal angle from which the system can recover if the initial position of the cart is $2.4\,m$ from either end of the track. The initial velocity of the cart and
		the initial angular velocity of the long pole are both $0$, and the maximal allowable force ($10\,N$) is applied constantly until the cart reaches the end of the track.
		The angles were tested from $1$\textdegree~to $36$\textdegree~in $1$\textdegree~increments. It is clear that the cart cannot push the long pole past the equilibrium point
		if the initial angle is $36$\textdegree~and the cart is more than $1.2\,m$ away from the centre of the track. This tendency becomes more pronounced
		as the initial position of the cart approaches the end of the track.}
	\label{fig:recoverability}
\end{figure}

This intuition was confirmed by a simulation of this experiment (Fig.~\ref{fig:recoverability}).
Looking at the plot, it is immediately apparent that the system cannot physically recover to a long-term stable condition
if the initial angle is $36$\textdegree~and the initial distance of the cart from the centre of the track is $1.2\,m$ or greater.
In other words, under these conditions the long pole will never go past the equilibrium point, even if the cart is
pushed \textit{continuously} with the maximal allowable force all the way to the end of the track.
Unsurprisingly, the situation worsens the closer the cart is to the end, which is alarming considering that this experiment does not even include
the effects of the initial velocity of the cart and initial angular velocity of the long pole.
Therefore, it seems that the set of initial conditions should fall within the recoverability region in order to determine the generalisation performance of the controller in a meaningful way.
Importance should also be given to the facts that the cart needs time to slow down to $0\,m/s$ \textit{before} reaching the end of the track,
and that the controller can never actually produce the maximal force (the $\tanh$ activation function approaches $\pm1$ asymptotically).
Furthermore, in generalisation tests in previous studies, the controller was tested for only 1000 time steps ($10\,s$ of simulation time) for each of the 625 conditions described above. However, 1000 time steps seems insufficient to claim that the controller has evolved a working strategy.

In light of the above considerations, the generalisation test in this study checked whether the controller was able to balance the poles when the cart's initial position was $-1.2$
to $1.2\,m$ away from the centre of the track ($1.2\,m$ away from the centre in either direction) in increments of $0.3\,m$, and the initial angle of the long pole was
$-15$\textdegree~to $15$\textdegree~in increments of $3$\textdegree.
The limit of $15$\textdegree~was chosen because the cart should be given enough time to slow down to $0\,m/s$ before reaching the end of the track from any of the starting positions.
These constraints were based on the results presented in Fig.~\ref{fig:recoverability}.
The remaining parameters were the same as those in Table~\ref{table:dpb}.
The controller was deemed a solution if it could balance the poles for \textit{each} combination of initial position and long pole angle for 200000 steps (more than $30\,min$ of simulation time).
50 experiments were performed, where an experiment was set to terminate after 1000 generations if no solution was found.

Cortex managed to generalise to all of the initial conditions in all 50 experiments with 27547 evaluations and 194 generations on average,
which is commendable given the recoverability region in Fig.~\ref{fig:recoverability}.
The average solution had four hidden neurons.
In comparison, NEAT required an average of 33184 evaluations to generalise to 286 of the 625 conditions \parencite{Stanley_Miikkulainen--2002}.
Unfortunately, direct comparison with previous generalisation results is infeasible because the evaluation methods differ.
In this regard, a more rigorous exploration of the recoverability region, including non-zero values for the initial speed and angular velocity for both poles,
should provide an even more complete set of initial conditions for generalisation testing.

\section{Conclusion}\label{sec:conclusion}
The approach to speciation and genotype representation proposed in this study, in combination with the accurate and reproducible network
evaluation procedure, is shown to be effective in rapidly finding solutions to common benchmark problems, and allows Cortex to converge
on a solution with reliability and speed competitive with existing platforms.

There are a number of points which remain to be addressed. First, the proposed encoding is \textit{direct}, which means that there is a
straightforward mapping of genotype to phenotype. As \textit{indirect} encoding has been shown to evolve solutions faster than direct encoding for the same problems
\parencite{Gauci_Stanley--2010,Risi_Stanley--2012}, it would be instructive to extend Cortex to support indirect encoding in order to
be able to compare it with existing indirect encoding NE schemes, such as (ES-)HyperNEAT.
In this regard, Cortex also supports mutation of the neuron activation function.
Although this idea has been considered in existing NE platforms,
it is still a somewhat rare concept which is employed, for example, in indirect encodings such as HyperNEAT \parencite{Huizinga_Clune_Mouret--2014}.
This mutation type introduces higher versatility and diversity in, for example, interactive NE applications,
and its potential use in an indirect encoding version of Cortex will be explored in future studies.

Furthermore, synaptic plasticity allows networks to adapt to the environment at \textit{runtime} by changing the weights in response to neuron activity.
In the experiments above, the weights are changed only by mutations, after which the network is evaluated with fixed weights.
It has been demonstrated that plasticity can play a positive role in allowing networks to adapt to their environment and to respond
more smoothly to sudden changes thereof \parencite{Risi_Stanley--2012b}.
Therefore, the effects of synaptic plasticity on the proposed framework will also be explored, particularly with respect to highly dynamic
tasks such as double pole balancing.

Lastly, preliminary experiments have indicated that the proposed genotype representation may also be beneficial for evolving the topology of spiking neural networks.
This direction of research has been poorly explored, likely due to the difficulties involved in evaluating spiking neural networks of arbitrary topology.
The Cortex software platform has been made available to promote research and development in this area.

\section*{Acknowledgment}

The authors would like to thank the three anonymous reviewers for their detailed and useful comments,
which were essential for improving the readability and understandability of the final revision of this paper.





%
\printbibliography

\end{document}